\documentclass{article} 
\usepackage{iclr2017,times}
\usepackage{hyperref}
\usepackage{url}

\usepackage[utf8]{inputenc}
\usepackage{amsmath}
\usepackage{amssymb}
\usepackage{csquotes}
\usepackage{graphicx} \graphicspath{{figures/}}
\usepackage{subfigure}
\usepackage{booktabs}

\title{Structured Sequence Modeling with \\ Graph Convolutional Recurrent Networks}

\author{Youngjoo Seo \\
EPFL, Switzerland \\
\texttt{youngjoo.seo@epfl.ch} \\
\And
Michaël Defferrard \\
EPFL, Switzerland \\
\texttt{michael.defferrard@epfl.ch} \\
\And
Pierre Vandergheynst \\
EPFL, Switzerland \\
\texttt{pierre.vandergheynst@epfl.ch} \\
\And
Xavier Bresson \\
EPFL, Switzerland \\
\texttt{xavier.bresson@epfl.ch} 
}

\DeclareMathOperator*{\diag}{diag}

\DeclareMathOperator*{\argmax}{arg\,max}

\newcommand{\R}{\mathbb{R}}
\newcommand{\bO}{\mathcal{O}}
\newcommand{\G}{\mathcal{G}}
\newcommand{\V}{\mathcal{V}}
\newcommand{\E}{\mathcal{E}}
\newcommand{\figref}[1]{Figure~\ref{fig:#1}}
\newcommand{\tabref}[1]{Table~\ref{tab:#1}}
\newcommand{\eqnref}[1]{(\ref{eqn:#1})}

\usepackage{color}

\begin{document}

\maketitle

\begin{abstract}
	This paper introduces Graph Convolutional Recurrent
	Network (GCRN), a deep learning model able to predict structured
	sequences of data. Precisely, GCRN is a generalization of classical recurrent neural networks (RNN) to
	data structured by an arbitrary graph. 
	Such structured sequences can represent series of frames in videos, spatio-temporal measurements on a network of
	sensors, or random walks on a vocabulary graph for natural language modeling.
	The proposed model combines convolutional neural networks (CNN) on graphs to identify spatial structures and RNN to find dynamic patterns. We study two possible architectures of GCRN, and apply the models to two practical
	problems: predicting moving MNIST data, and modeling natural language with the Penn
	Treebank dataset. Experiments show that exploiting simultaneously graph spatial and dynamic information about data can improve both precision and learning speed.
\end{abstract}

\section{Introduction}


Many real-world data can be cast as structured sequences, with spatio-temporal
sequences being a special case. A well-studied example of spatio-temporal data are videos, where
succeeding frames share temporal and spatial structures. Many works, such as \citet{cnnlstm1, cnnlstm2, cnnlstm3},  leveraged a combination of CNN and RNN to exploit such spatial and
temporal regularities. Their models are able to process possibly time-varying
visual inputs for variable-length prediction. These neural network architectures consist of combining
a CNN for visual feature extraction followed by a RNN for sequence learning. Such
architectures have been successfully used for video activity recognition, image
captioning and video description.

More recently, interest has grown in properly fusing the CNN and RNN models for
spatio-temporal sequence modeling. Inspired by language modeling,
\citet{video_language_model} proposed a model to represent complex deformations
and motion patterns by discovering both spatial and temporal correlations. They showed that prediction of the next video frame and
interpolation of intermediate frames can be achieved by building a RNN-based
language model on the visual words obtained by quantizing the image patches.
Their highest-performing model, recursive CNN (rCNN), uses convolutions for both
 inputs and states. \citet{convlstm}  then proposed the convolutional
LSTM network (convLSTM), a recurrent model for spatio-temporal sequence
modeling which uses 2D-grid convolution to leverage the spatial correlations in
input data. They successfully applied their model to the prediction of the
evolution of radar echo maps for precipitation nowcasting.



The spatial structure of many important problems may however not be as simple as regular grids. For instance, 
the data measured from meteorological stations lie on a irregular grid, i.e. a network of heterogeneous spatial
distribution of stations. More challenging, the spatial structure of data may not even be spatial, as it is the case for social or biological networks. Eventually, the interpretation that sentences can be regarded as random walks on vocabulary graphs,
a view popularized by \cite{word2vec}, allows us to cast language analysis problems as graph-structured sequence models.

\begin{figure}[t]
	\begin{minipage}[t]{0.59\linewidth}
		\includegraphics[width=\linewidth]{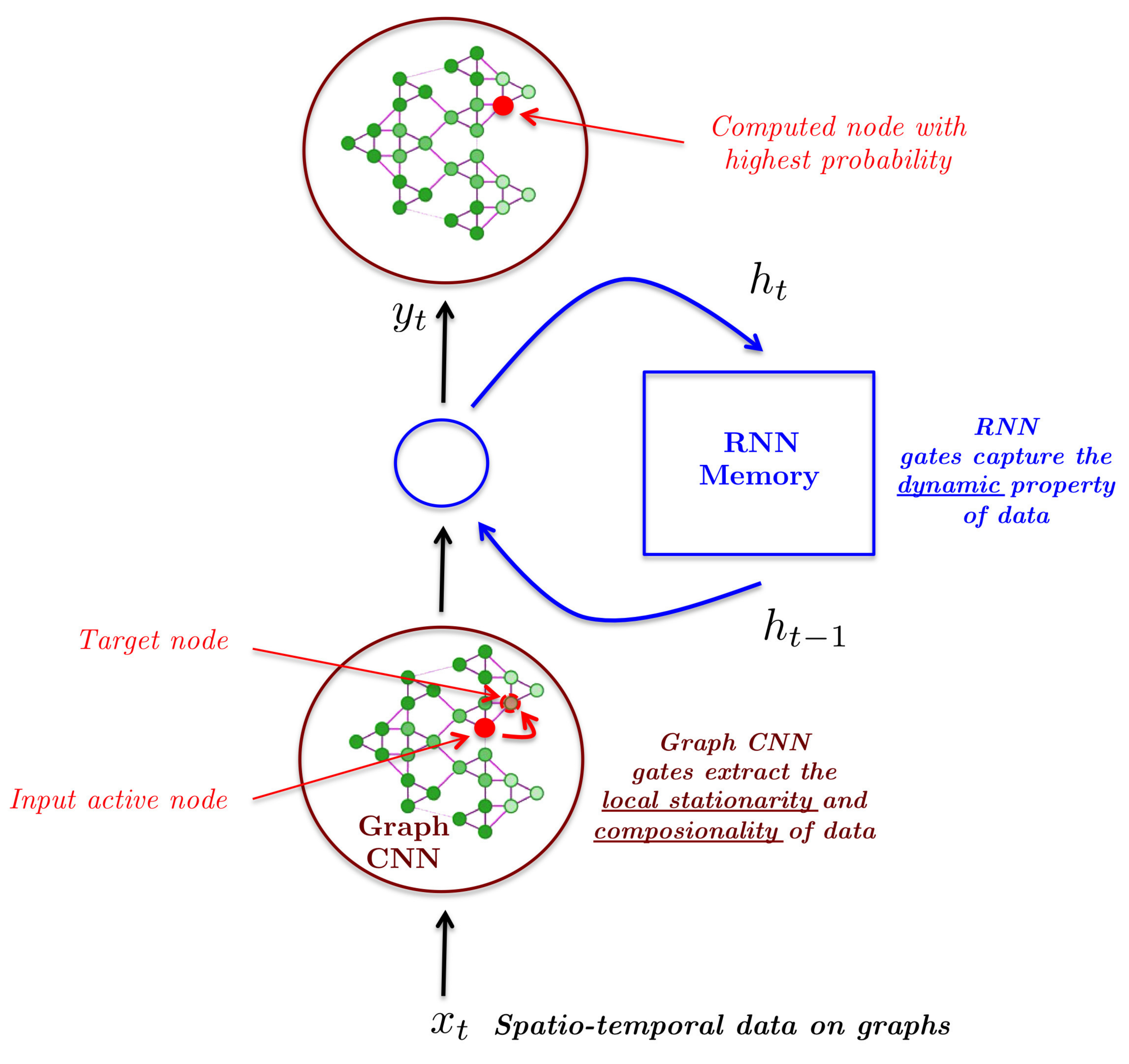}
		\caption{Illustration of the proposed GCRN model for spatio-temporal prediction of graph-structured data. The technique combines at the same time CNN on graphs and RNN. RNN can be easily exchanged with LSTM or GRU networks.}
		\label{fig:gcrn}
	\end{minipage} \hfill
	\begin{minipage}[t]{0.37\linewidth}
		\includegraphics[width=\linewidth]{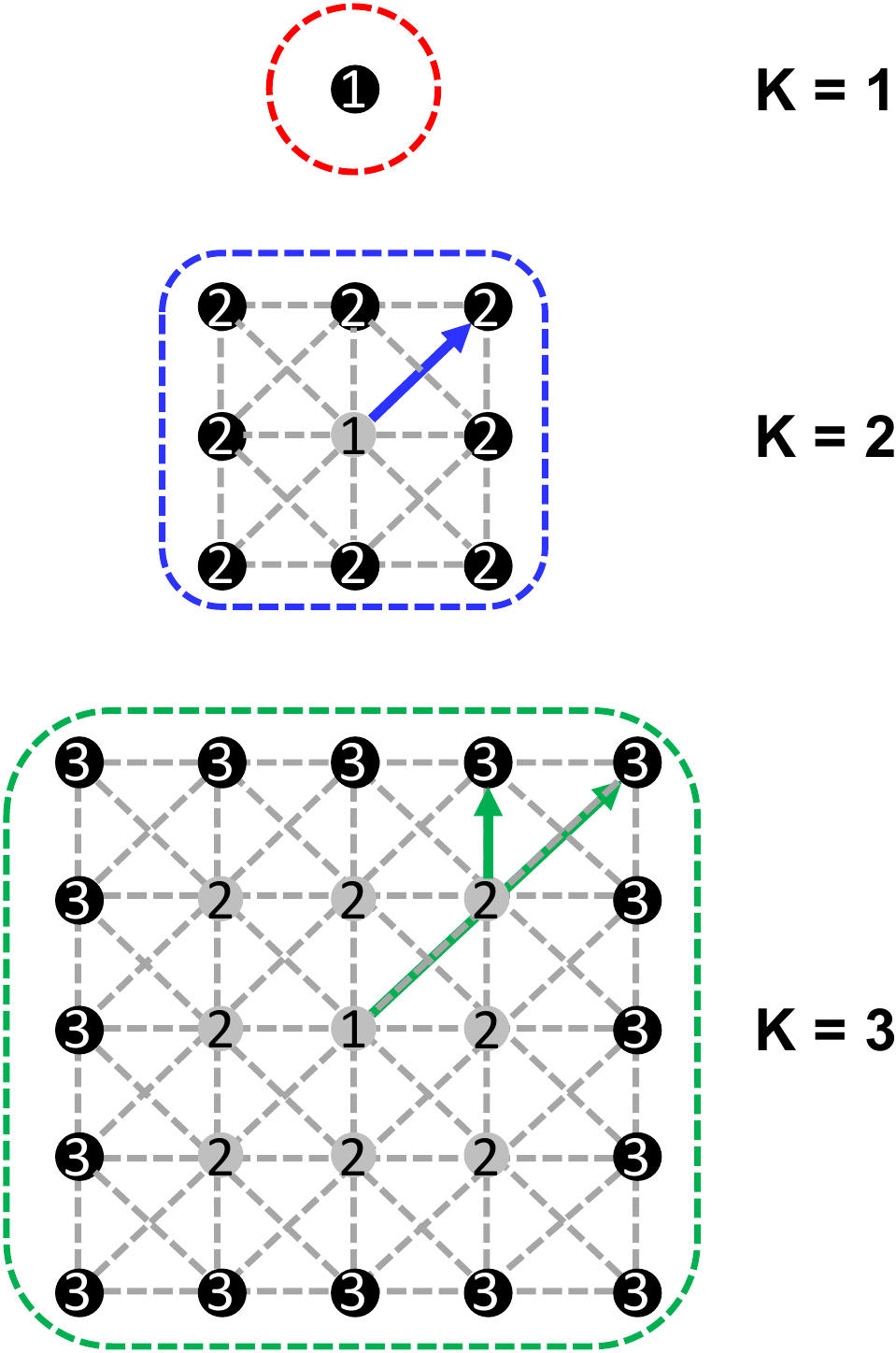}
		\caption{Illustration of the neighborhood on an 8-nearest-neighbor grid graph. Isotropic spectral filters of support $K$ have access to nodes at most at $K-1$ hops.}
		\label{fig:reach}
	\end{minipage}
\end{figure}

This work leverages on the recent models of
\cite{gcnn,video_language_model,convlstm} to design the GCRN model for modeling
and predicting time-varying graph-based data. The core idea is to merge CNN for
graph-structured data and RNN to identify simultaneously meaningful spatial
structures and dynamic patterns. A generic illustration of the proposed GCRN
architecture is given by \figref{gcrn}.

\section{Preliminaries}


\subsection{Structured Sequence Modeling}

Sequence modeling is the problem of predicting the most likely future
length-$K$ sequence given the previous $J$ observations:
\begin{equation} \label{eqn:seq}
	\hat{x}_{t+1}, \ldots, \hat{x}_{t+K} =
	\argmax_{x_{t+1}, \ldots, x_{t+K}}
	P(x_{t+1}, \ldots, x_{t+K} | x_{t-J+1}, \ldots, x_t),
\end{equation}
where $x_t \in \mathbf{D}$ is an observation at time $t$ and $\mathbf{D}$
denotes the domain of the observed features. The archetypal application being
the $n$-gram language model (with $n = J+1$), where $P(x_{t+1} | x_{t-J+1},
\ldots, x_t)$ models the probability of word $x_{t+1}$ to appear conditioned on
the past $J$ words in the sentence \citep{seq_graves}.

In this paper, we are interested in special structured sequences, i.e. sequences where features of the observations $x_t$ are not independent but linked by pairwise relationships. Such relationships are universally modeled by weighted graphs. 

Data $x_t$ can be viewed as a graph signal, i.e.
a signal defined on an undirected and weighted graph $\G = (\V, \E, A)$, where
$\V$ is a finite set of $|\V| = n$ vertices, $\E$ is a set of edges and $A \in
\R^{n \times n}$ is a weighted adjacency matrix encoding the connection weight
between two vertices. A signal $x_t: \V \rightarrow \R^{d_x}$ defined on the
nodes of the graph may be regarded as a matrix $x_t \in \R^{n \times d_x}$
whose column $i$ is the $d_x$-dimensional value of $x_t$ at the $i^{th}$ node.
While the number of free variables in a structured sequence of length $K$ is in
principle $\bO(n^K {d_x}^K)$, we seek to exploit the structure of the space
of possible predictions to reduce the dimensionality and hence make those
problems more tractable.

\subsection{Long Short-Term Memory}


A special class of recurrent neural networks (RNN) that prevents the gradient from vanishing
too quickly is the popular long short-term memory (LSTM) introduced by \citet{lstm}. This architecture has proven stable and powerful for modeling long-range dependencies in various general-purpose sequence modeling tasks
\citep{seq_graves, moving_mnist, seq2seq}. A fully-connected LSTM (FC-LSTM) may
be seen as a multivariate version of LSTM where the input $x_t \in \R^{d_x}$,
cell output $h_t \in [-1,1]^{d_h}$ and states $c_t \in \R^{d_h}$ are all
vectors. In this paper, we follow the FC-LSTM formulation of
\citet{seq_graves}, that is:
\begin{align} \label{eqn:lstm_fc}
\begin{split}
	i &= \sigma(W_{xi} x_t + W_{hi} h_{t-1} + w_{ci} \odot c_{t-1} + b_i), \\
	f &= \sigma(W_{xf} x_t + W_{hf} h_{t-1} + w_{cf} \odot c_{t-1} + b_f), \\
	c_t &= f_t \odot c_{t-1} + i_t \odot \tanh(W_{xc} x_t + W_{hc} h_{t-1} + b_c), \\
	o &= \sigma(W_{xo} x_t + W_{ho} h_{t-1} + w_{co} \odot c_t + b_o), \\
	h_t &= o \odot \tanh(c_t),
\end{split}
\end{align}
where $\odot$ denotes the Hadamard product, $\sigma(\cdot)$ the sigmoid
function $\sigma(x) = 1 / (1+e^{-x})$ and $i, f, o \in [0,1]^{d_h}$ are the
input, forget and output gates. The weights $W_{x\cdot} \in \R^{d_h \times
d_x}$, $W_{h\cdot} \in \R^{d_h \times d_h}$, $w_{c\cdot} \in \R^{d_h}$ and
biases $b_i, b_f, b_c, b_o \in \R^{d_h}$ are the model parameters.\footnote{A
practical trick is to initialize the biases $b_i$, $b_f$ and $b_o$ to one such
that the gates are initially open.} Such a model is called fully-connected
because the dense matrices $W_{x\cdot}$ and $W_{h\cdot}$ linearly combine all
the components of $x$ and $h$. The optional peephole connections $w_{c\cdot}
\odot c_t$, introduced by \citet{peephole}, have been found to improve
performance on certain tasks.


\subsection{Convolutional Neural Networks on Graphs} \label{sec:graphconv}

Generalizing convolutional neural networks (CNNs) to arbitrary graphs is a
recent area of interest. Two approaches have been explored in the literature: (i) a
generalization of the spatial definition of a convolution \citep{gcnn_masci,
gcnn_niepert} and (ii), a multiplication in the graph Fourier domain by the way
of the convolution theorem \citep{gcnn_bruna, gcnn}. \citet{gcnn_masci}
introduced a spatial generalization of CNNs to 3D meshes. The authors used
geodesic polar coordinates to define convolution operations on mesh patches,
and formulated a deep learning architecture which allows comparison across
different meshes.  Hence, this method is tailored to manifolds and is not directly
generalizable to arbitrary graphs. \citet{gcnn_niepert} proposed a spatial
approach which may be decomposed in three steps: (i) select a node, (ii)
construct its neighborhood and (iii) normalize the selected sub-graph, i.e.
order the neighboring nodes. The extracted patches are then fed into a
conventional 1D Euclidean CNN. As graphs generally do not possess a natural ordering
(temporal, spatial or otherwise), a labeling procedure should be used to impose
it. \citet{gcnn_bruna} were the first to introduce the spectral framework
described below in the context of graph CNNs. The major drawback of this
method is its $\bO(n^2)$ complexity, which was overcome with the technique of \citet{gcnn}, which offers a linear complexity $\bO(|\E|)$ and provides strictly localized
filters. \citet{gcnn_kipf} took a first-order approximation of the spectral
filters proposed by \citet{gcnn} and successfully used it for semi-supervised
classification of nodes. While we focus on the framework introduced by
\citet{gcnn}, the proposed model is agnostic to the choice of the graph
convolution operator $\ast_\G$. 

As it is difficult to express a meaningful translation operator in the vertex
domain \citep{gcnn_bruna, gcnn_niepert}, \citet{gcnn} chose a spectral
formulation for the convolution operator on graph $\ast_\G$. By this
definition, a graph signal $x \in \R^{n \times d_x}$ is filtered by a non-parametric kernel
$g_\theta(\Lambda) = \diag(\theta)$, where $\theta \in \R^n$ is a vector of
Fourier coefficients, as
\begin{equation} \label{eqn:graph_conv}
	y = g_\theta \ast_\G x = g_\theta(L) x =
	g_\theta(U \Lambda U^T) x = U g_\theta(\Lambda) U^T x \in \R^{n \times d_x},
\end{equation}
where $U \in \R^{n \times n}$ is the matrix of eigenvectors and $\Lambda \in
\R^{n \times n}$ the diagonal matrix of eigenvalues of the normalized graph
Laplacian $L = I_n - D^{-1/2} A D^{-1/2} = U \Lambda U^T \in \R^{n \times n}$,
where $I_n$ is the identity matrix and $D \in \R^{n \times n}$ is the diagonal
degree matrix with $D_{ii} = \sum_j A_{ij}$ \citep{chung}.  Note that the
signal $x$ is filtered by $g_\theta$ with an element-wise multiplication of its
graph Fourier transform $U^T x$ with $g_\theta$ \citep{gsp}. Evaluating
\eqnref{graph_conv} is however expensive, as the multiplication with $U$ is
$\bO(n^2)$. Furthermore, computing the eigendecomposition of $L$ might be
prohibitively expensive for large graphs. To circumvent this problem,
\cite{gcnn} parametrizes $g_\theta$ as a truncated expansion, up to order
$K-1$, of Chebyshev polynomials $T_k$ such that
\begin{equation} \label{eqn:filt_cheby}
	g_\theta(\Lambda) = \sum_{k=0}^{K-1} \theta_k T_k(\tilde{\Lambda}),
\end{equation}
where the parameter $\theta \in \R^K$ is a vector of Chebyshev coefficients and
$T_k(\tilde{\Lambda}) \in \R^{n \times n}$ is the Chebyshev polynomial of order
$k$ evaluated at $\tilde{\Lambda} = 2 \Lambda / \lambda_{max} - I_n$. The
graph filtering operation can then be written as
\begin{equation} \label{eqn:graph_conv_cheby}
	y = g_\theta \ast_\G x = g_\theta(L) x = \sum_{k=0}^{K-1} \theta_k T_k(\tilde{L}) x,
\end{equation}
where $T_k(\tilde{L}) \in \R^{n \times n}$ is the Chebyshev polynomial of order
$k$ evaluated at the scaled Laplacian $\tilde{L} = 2 L / \lambda_{max} - I_n$.
Using the stable recurrence relation $T_k(x) = 2x T_{k-1}(x) - T_{k-2}(x)$ with
$T_0 = 1$ and $T_1 = x$, one can evaluate \eqnref{graph_conv_cheby} in
$\bO(K|\E|)$ operations, i.e. linearly with the number of edges. Note that as
the filtering operation \eqnref{graph_conv_cheby} is an order $K$ polynomial of
the Laplacian, it is $K$-localized and depends only on nodes that are at
maximum $K$ hops away from the central node, the $K$-neighborhood. The reader
is referred to \cite{gcnn} for details and an in-depth discussion.

\section{Related Works}


\citet{convlstm} introduced a model for regular grid-structured sequences, which can be
seen as a special case of the proposed model where the graph is an image grid where the
nodes are well ordered. Their model is essentially the classical FC-LSTM
\eqnref{lstm_fc} where the multiplications by dense matrices $W$ have been
replaced by convolutions with kernels $W$:
\begin{align} \label{eqn:lstm_conv}
\begin{split}
	i &= \sigma(W_{xi} \ast x_t + W_{hi} \ast h_{t-1} +
	            w_{ci} \odot c_{t-1} + b_i), \\
	f &= \sigma(W_{xf} \ast x_t + W_{hf} \ast h_{t-1} +
	            w_{cf} \odot c_{t-1} + b_f), \\
	c_t &= f_t \odot c_{t-1} + i_t \odot
	       \tanh(W_{xc} \ast x_t + W_{hc} \ast h_{t-1} + b_c), \\
	o &= \sigma(W_{xo} \ast x_t + W_{ho} \ast h_{t-1} +
	            w_{co} \odot c_t + b_o), \\
	h_t &= o \odot \tanh(c_t),
\end{split}
\end{align}
where $\ast$ denotes the 2D convolution by a set of kernels. In their setting,
the input tensor $x_t \in \R^{n_r \times n_c \times d_x}$ is the observation of
$d_x$ measurements at time $t$ of a dynamical system over a spatial region
represented by a grid of $n_r$ rows and $n_c$ columns. The model holds
spatially distributed hidden and cell states of size $d_h$ given by the tensors
$c_t, h_t \in \R^{n_r \times n_c \times d_h}$. The size $m$ of the
convolutional kernels $W_{h\cdot} \in \R^{m \times m \times d_h \times d_h}$
and $W_{x\cdot} \in \R^{m \times m \times d_h \times d_x}$ determines the
number of parameters, which is independent of the grid size $n_r \times n_c$.
Earlier, \citet{video_language_model} proposed a similar RNN variation which
uses convolutional layers instead of fully connected layers. The hidden state
at time $t$ is given by
\begin{equation}
	h_t = \tanh(\sigma(W_{x2} \ast \sigma(W_{x1} \ast x_t)) + \sigma(W_h \ast h_{t-1})),
\end{equation}
where the convolutional kernels $W_h \in \R^{d_h \times d_h}$ are restricted to
filters of size 1x1 (effectively a fully connected layer shared across all
spatial locations).

Observing that natural language exhibits syntactic properties that naturally
combine words into phrases, \citet{treelstm} proposed a model for
tree-structured topologies, where each LSTM has access to the states of its
children. They obtained state-of-the-art results on semantic relatedness and
sentiment classification. \citet{graphlstm} followed up and proposed a variant
on graphs. Their sophisticated network architecture obtained state-of-the-art
results for semantic object parsing on four datasets. In those models, the
states are gathered from the neighborhood by way of a weighted sum with
trainable weight matrices. Those weights are however not shared across the
graph, which would otherwise have required some ordering of the nodes, alike
any other spatial definition of graph convolution. Moreover, their formulations
are limited to the one-neighborhood of the current node, with equal weight
given to each neighbor.

Motivated by spatio-temporal problems like modeling human motion and object
interactions, \citet{structuralrnn} developed a method to cast a
spatio-temporal graph as a rich RNN mixture which essentially associates a RNN
to each node and edge. Again, the communication is limited to directly
connected nodes and edges.


The closest model to our work is probably the one proposed by \citet{ggsnn_li},
which showed stat-of-the-art performance on a problem from program
verification. Whereas they use the iterative procedure of the Graph Neural
Networks (GNNs) model introduced by \citet{gnn_scarcelli} to propagate node
representations until convergence, we instead use the graph CNN introduced by
\citet{gcnn} to diffuse information across the nodes. While their motivations
are quite different, those models are related by the fact that a spectral
filter defined as a polynomial of order $K$ can be implemented as a $K$-layer
GNN.\footnote{The basic idea is to set the transition function as a diffusion
and the output function such as to realize the polynomial recurrence, then
stack $K$ of those. See \citet{gcnn} for details.}

\section{Proposed GCRN Models}
\label{sec:our_models}


We propose two GCRN architectures that are quite natural, and investigate their
performances in real-world applications in Section \ref{experiments}.

\paragraph{Model 1.} The most straightforward definition is to stack a graph
CNN, defined as \eqnref{graph_conv_cheby}, for feature extraction and an LSTM,
defined as \eqnref{lstm_fc}, for sequence learning:
\begin{align} \label{eqn:lstm_graph_v1}
\begin{split}
	x_t^{\textrm{CNN}} &=  \textrm{CNN}_\G(x_t)\\
	i &= \sigma(W_{xi} x_t^{\textrm{CNN}} + W_{hi} h_{t-1}  +
	            w_{ci} \odot c_{t-1} + b_i), \\
	f &= \sigma(W_{xf} x_t^{\textrm{CNN}} + W_{hf} h_{t-1} + w_{cf} \odot c_{t-1} + b_f), \\
	c_t &= f_t \odot c_{t-1} + i_t \odot \tanh(W_{xc} x_t^{\textrm{CNN}} + W_{hc} h_{t-1} + b_c), \\
	o &= \sigma(W_{xo} x_t^{\textrm{CNN}} + W_{ho} h_{t-1} + w_{co} \odot c_t + b_o), \\
	h_t &= o \odot \tanh(c_t).
\end{split}
\end{align}
In that setting, the input matrix $x_t \in \R^{n \times d_x}$ may represent the
observation of $d_x$ measurements at time $t$ of a dynamical system over a
network whose organization is given by a graph $\G$. $x_t^{\textrm{CNN}}$ is the output of the graph CNN gate. For a proof of concept, we simply choose here $x_t^{\textrm{CNN}} = W^{\textrm{CNN}} \ast_\G x_t$, where $W^{\textrm{CNN}} \in \R^{K \times d_x \times d_x}$ are the Chebyshev coefficients  for the graph convolutional kernels of support $K$. The model also holds spatially distributed hidden and cell states of size $d_h$ given by the matrices $c_t, h_t \in \R^{n \times d_h}$. Peepholes are controlled by $w_{c\cdot} \in \R^{n \times d_h}$. The weights $W_{h\cdot} \in
\R^{ d_h \times d_h}$ and $W_{x\cdot} \in \R^{d_h \times d_x}$ are the parameters of the fully connected layers. An architecture such as \eqnref{lstm_graph_v1} may be enough to capture the data distribution by exploiting local stationarity and compositionality properties as well as the dynamic properties.

\paragraph{Model 2.} To generalize the convLSTM model \eqnref{lstm_conv} to
graphs we replace the Euclidean 2D convolution $\ast$ by the graph convolution
$\ast_\G$:
\begin{align} \label{eqn:lstm_graph_v2}
\begin{split}
	i &= \sigma(W_{xi} \ast_\G x_t + W_{hi} \ast_\G h_{t-1} +
	            w_{ci} \odot c_{t-1} + b_i), \\
	f &= \sigma(W_{xf} \ast_\G x_t + W_{hf} \ast_\G h_{t-1} +
	            w_{cf} \odot c_{t-1} + b_f), \\
	c_t &= f_t \odot c_{t-1} + i_t \odot
	       \tanh(W_{xc} \ast_\G x_t + W_{hc} \ast_\G h_{t-1} + b_c), \\
	o &= \sigma(W_{xo} \ast_\G x_t + W_{ho} \ast_\G h_{t-1} +
	            w_{co} \odot c_t + b_o), \\
	h_t &= o \odot \tanh(c_t).
\end{split}
\end{align}
In that setting, the support $K$ of the graph convolutional kernels defined by
the Chebyshev coefficients $W_{h\cdot} \in \R^{K \times d_h \times d_h}$ and
$W_{x\cdot} \in \R^{K \times d_h \times d_x}$ determines the number of
parameters, which is independent of the number of nodes $n$.  To keep the
notation simple, we write $W_{xi} \ast_\G x_t$ to mean a graph convolution of
$x_t$ with $d_h d_x$ filters which are functions of the graph Laplacian $L$
parametrized by $K$ Chebyshev coefficients, as noted in \eqnref{filt_cheby} and
\eqnref{graph_conv_cheby}. In a distributed computing setting, $K$ controls
the communication overhead, i.e. the number of nodes any given node $i$ should
exchange with in order to compute its local states.

The proposed blend of RNNs and graph CNNs is not limited to LSTMs and is
straightforward to apply to any kind of recursive networks. For example, a
vanilla RNN $h_t = \tanh(W_x x_{t} + W_h h_{t-1})$ would be modified as
\begin{equation} \label{eqn:vrnn_graph}
	h_t = \tanh(W_x \ast_\G x_t + W_h \ast_\G h_{t-1}),
\end{equation}
and a Gated Recurrent Unit (GRU) \citep{gru} as
\begin{align} \label{eqn:gru_graph}
\begin{split}
	z &= \sigma(W_{xz} \ast_\G x_t + W_{hz} \ast_\G h_{t-1}), \\
	r &= \sigma(W_{xr} \ast_\G x_t + W_{hr} \ast_\G h_{t-1}), \\
	\tilde{h} &= \tanh(W_{xh} \ast_\G x_t + W_{hh} \ast_\G (r \odot h_{t-1})), \\
	h_t &= z \odot h_{t-1} + (1-z) \odot \tilde{h}.
\end{split}
\end{align}

As demonstrated by \citet{convlstm}, structure-aware LSTM cells can be stacked
and used as sequence-to-sequence models using an architecture composed of an
encoder, which processes the input sequence, and a decoder, which generates an
output sequence. A standard practice for machine translation using RNNs
\citep{gru, seq2seq}.

\section{Experiments} \label{experiments}

\subsection{Spatio-Temporal Sequence Modeling on Moving-MNIST}

For this synthetic experiment, we use the moving-MNIST dataset generated by
\citet{convlstm}. All sequences are 20 frames long (10 frames as input and 10
frames for prediction) and contain two handwritten digits bouncing inside a $64
\times 64$ patch. Following their experimental setup, all models are trained by
minimizing the binary cross-entropy loss using back-propagation through time
(BPTT) and RMSProp with a learning rate of $10^{-3}$ and a decay rate of 0.9.
We choose the best model with early-stopping on validation set. All
implementations are based on their Theano code and
dataset.\footnote{\url{http://www.wanghao.in/code/SPARNN-release.zip}} The
adjacency matrix $A$ is constructed as a k-nearest-neighbor (knn) graph with
Euclidean distance and Gaussian kernel between pixel locations. For a fair
comparison with \citet{convlstm} defined in \eqnref{lstm_conv}, all GCRN
experiments are conducted with Model 2 defined in \eqnref{lstm_graph_v2}, which
is the same architecture with the 2D convolution $\ast$ replaced by a graph
convolution $\ast_\G$. To further explore the impact of the isotropic property
of our filters, we generated a variant of the moving MNIST dataset where digits
are also rotating (see \figref{mMNIST_img}).

\begin{table}
	\centering
	{\small
	\begin{tabular}{lccccccc}
		\toprule
		Architecture & Structure & Filter size & Parameters & Runtime & Test(w/o Rot) & Test(Rot) \\
		\midrule
		FC-LSTM & N/A & N/A & 142,667,776 & N/A & 4832 & - \\
		LSTM+CNN & N/A & $5 \times 5$ & 13,524,496 & 2.10 & 3851 & 4339 \\
		LSTM+CNN & N/A & $9 \times 9$ & 43,802,128 & 6.10 & 3903 & 4208 \\

		LSTM+GCNN & $knn=8$ & $K=3$ & 1,629,712 & 0.82 & 3866 & 4367 \\
		LSTM+GCNN & $knn=8$ & $K=5$ & 2,711,056 & 1.24 & 3495 & 3932 \\
		LSTM+GCNN & $knn=8$ & $K=7$ & 3,792,400 & 1.61 & {\bf 3400} & {\bf 3803} \\
		LSTM+GCNN & $knn=8$ & $K=9$ & 4,873,744 & 2.15 & 3395 & 3814 \\
		
		LSTM+GCNN & $knn=4$ & $K=7$ & 3,792,400 & 1.61 & 3446 & 3844 \\
		LSTM+GCNN & $knn=16$ & $K=7$ & 3,792,400 & 1.61 & 3578 & 3963 \\
		\bottomrule
	\end{tabular}
	}
	\caption{Comparison between models. Runtime is the time spent per each mini-batch in seconds. Test cross-entropies correspond to moving MNIST, and rotating and moving MNIST. LSTM+GCNN is Model 2 defined in \eqnref{lstm_graph_v2}. Cross-entropy of FC-LSTM is taken from \cite{convlstm}.}
	\label{tab:moving_mnist}
\end{table}

\begin{figure}[t]
	\centering
	\subfigure{\includegraphics[width=0.48\linewidth]{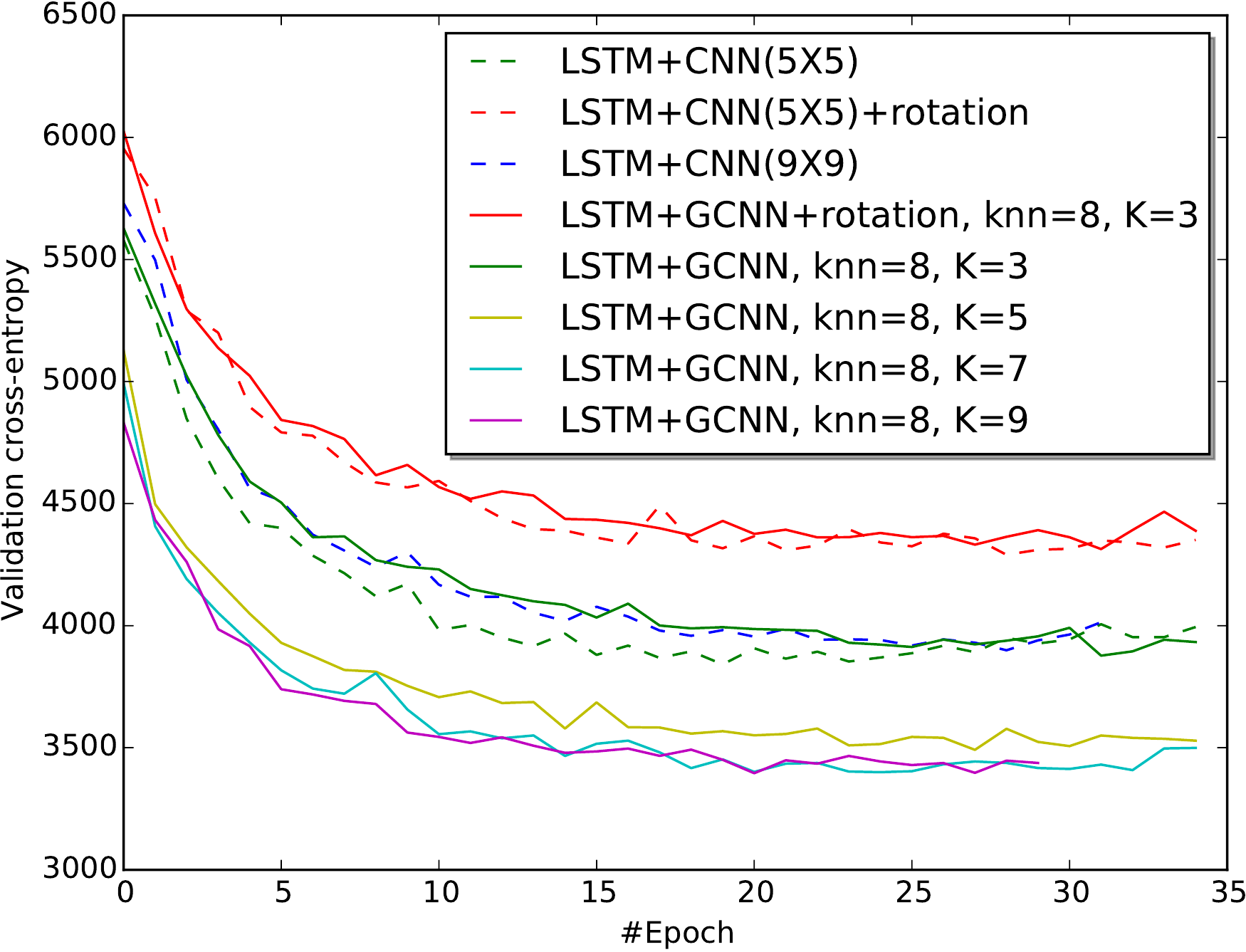}}
	\hfill
	\subfigure{\includegraphics[width=0.48\linewidth]{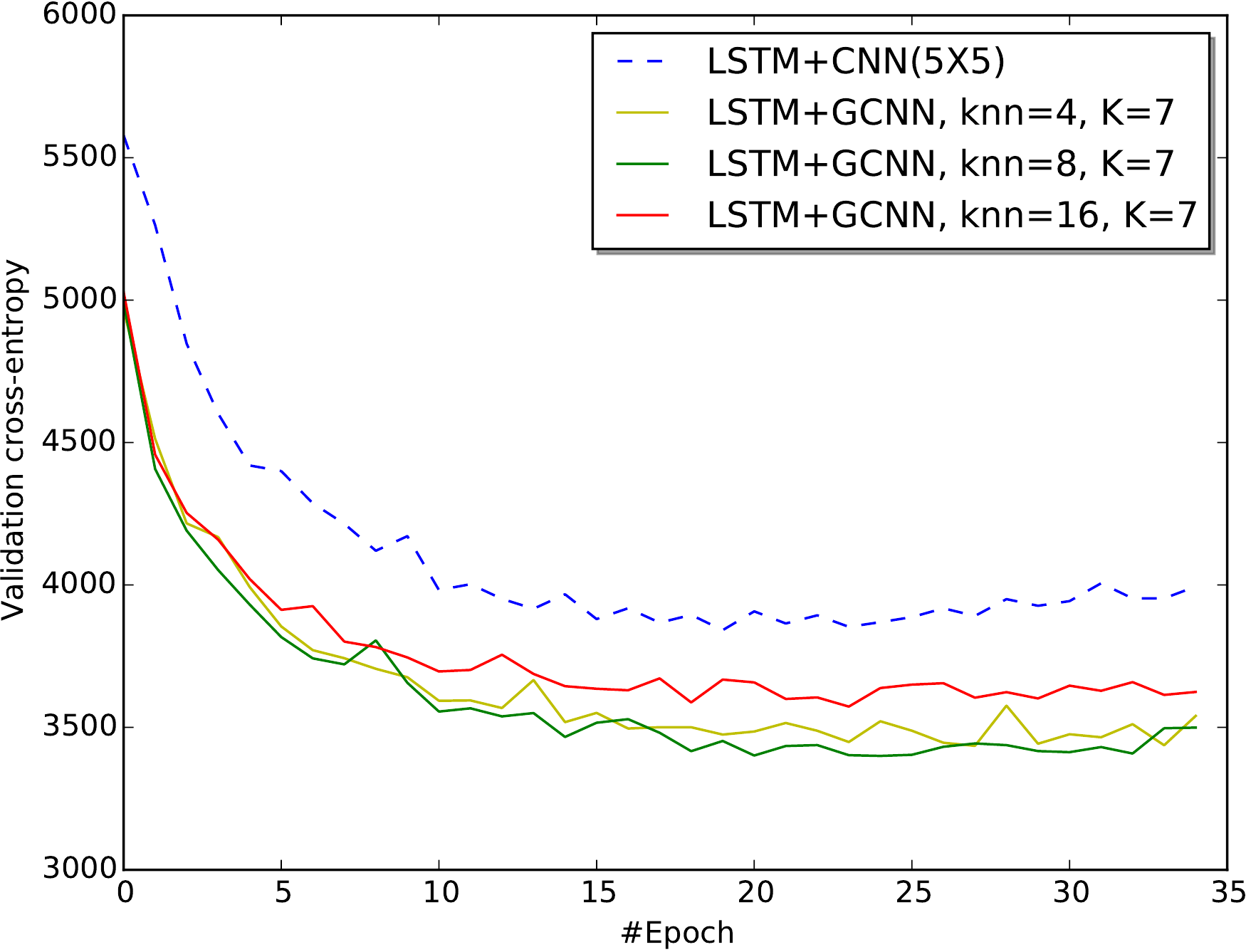}}
	\caption{Cross-entropy on validation set: Left: performance of graph CNN with various filter support $K$. Right: performance w.r.t. graph construction.}
	\label{fig:mMNIST_graph}
\end{figure}

\begin{figure}[t]
	\centering
	\subfigure{\includegraphics[width=0.48\linewidth]{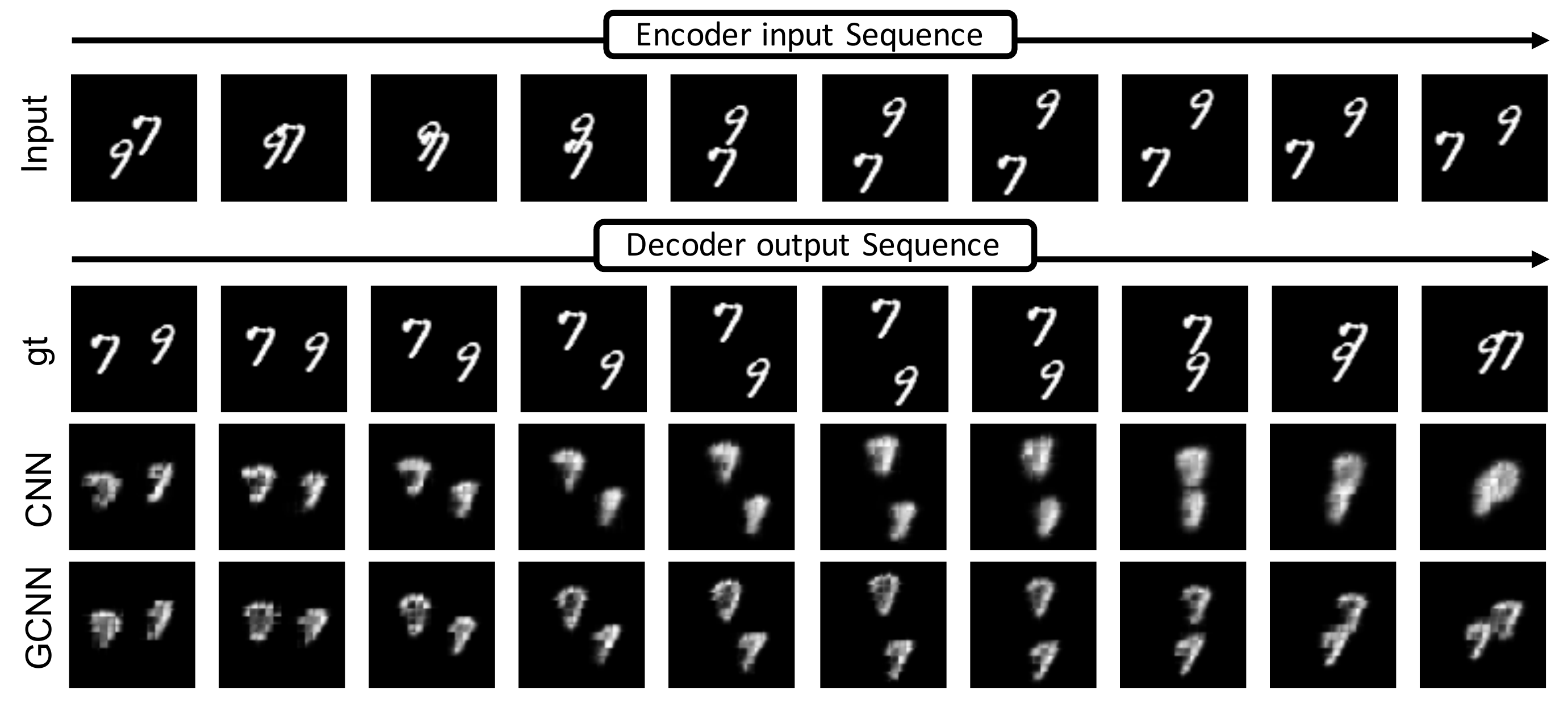}}
	\hfill
	\subfigure{\includegraphics[width=0.48\linewidth]{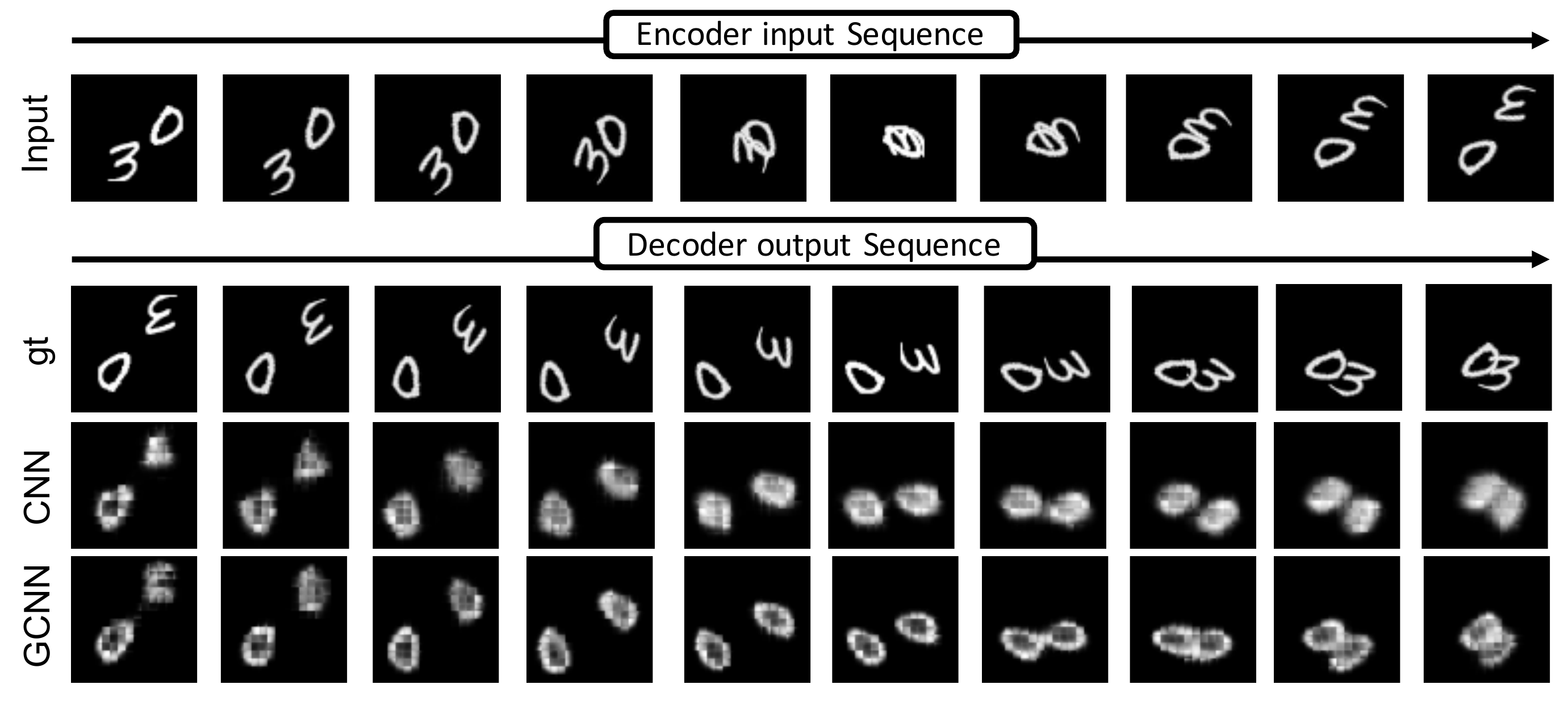}}
	\caption{Qualitative results for moving MNIST, and rotating and moving MNIST. First row is the input sequence, second the ground truth, and third and fourth are the predictions of the LSTM+CNN($5\times5$) and LSTM+GCNN($knn=8, K=7$).}
	\label{fig:mMNIST_img}
\end{figure}

\tabref{moving_mnist} shows the performance of various models: (i) the baseline
FC-LSTM from \citet{convlstm}, (ii) the 1-layer LSTM+CNN from \citet{convlstm}
with different filter sizes, and (iii) the proposed LSTM+graph CNN(GCNN)
defined in \eqnref{lstm_graph_v2} with different supports $K$. These results
show the ability of the proposed method to capture spatio-temporal structures.
Perhaps surprisingly, GCNNs can offer better performance than regular CNNs,
even when the domain is a 2D grid and the data is images, the problem CNNs were
initially developed for. The explanation is to be found in the differences
between 2D filters and spectral graph filters. While a spectral filter of
support $K=3$ corresponds to the reach of a patch of size $5 \times 5$ (see
\figref{reach}), the difference resides in the isotropic nature of the former
and the number of parameters: $K=3$ for the former and $5^2=25$ for the later.
\tabref{moving_mnist} indeed shows that LSTM+CNN($5\times5$) rivals LSTM+GCNN
with $K=3$. However, when increasing the filter size to $9\times9$ or $K=5$,
the GCNN variant clearly outperforms the CNN variant. This experiment
demonstrates that graph spectral filters can obtain superior performance on
regular domains with much less parameters thanks to their isotropic nature, a
controversial property. Indeed, as the nodes are not ordered, there is no
notion of an edge going up, down, on the right or on the left. All edges are
treated equally, inducing some sort of rotation invariance. Additionally,
\tabref{moving_mnist} shows that the computational complexity of each model is
linear with the filter size, and \figref{mMNIST_graph} shows the learning
dynamic of some of the models.

\subsection{Natural Language Modeling on Penn Treebank}

The Penn Treebank dataset has 1,036,580 words. It was pre-processed in \cite{zaremba2014recurrent} and split\footnote{\url{https://github.com/wojzaremba/lstm}} into a training set of 929k words, a validation set of 73k words, and a test set of 82k words. The size of the vocabulary of this corpus is 10,000. We use the gensim library\footnote{\url{https://radimrehurek.com/gensim/models/word2vec.html}} to compute a word2vec model \citep{word2vec} for embedding the words of the dictionary in a 200-dimensional space. Then we build the adjacency matrix of the word embedding using a 4-nearest neighbor graph with cosine distance. Figure \ref{fig2} presents the computed adjacency matrix, and its 3D visualization. We used the hyperparameters of the small configuration given by the code\footnotemark[6] based on \cite{zaremba2014recurrent}: the size of the data mini-batch is 20, the number of temporal steps to unroll is 20, the dimension of the hidden state is 200. The global learning rate is 1.0 and the norm of the gradient is bounded by 5. The learning decay function is selected to be $0.5^{\max(0,\textrm{\#epoch}-4)}$. All experiments have 13 epochs, and dropout value is 0.75. For \cite{zaremba2014recurrent}, the input representation $x_t$ can be either the 200-dim embedding vector of the word, or the 10,000-dim one-hot representation of the word. For our models, the input representation is a one-hot representation of the word. This choice allows us to use the graph structure of the words.

\begin{figure}[t]
	\centering
	\subfigure{\includegraphics[width=0.48\linewidth]{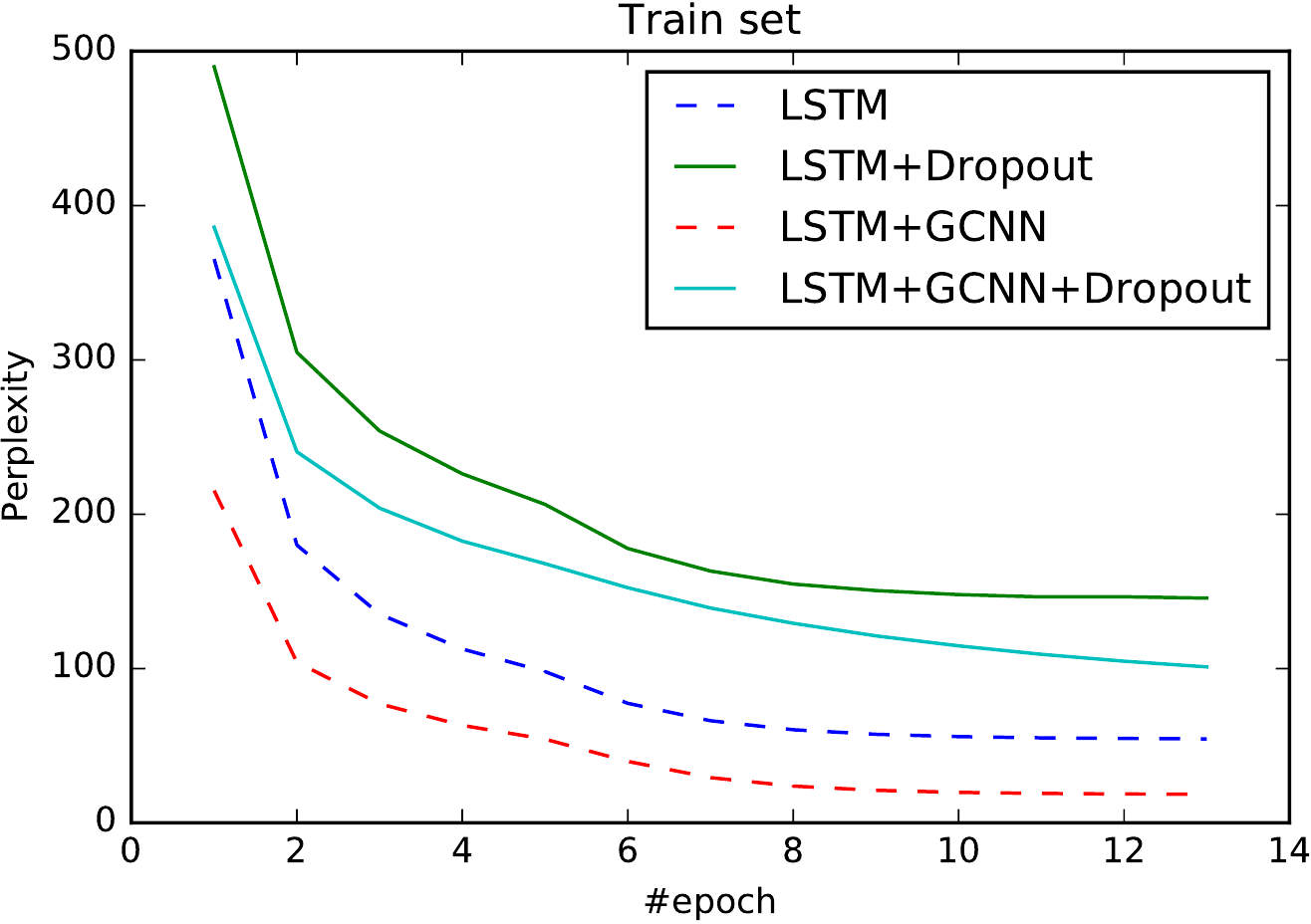}}
	\hfill
	\subfigure{\includegraphics[width=0.48\linewidth]{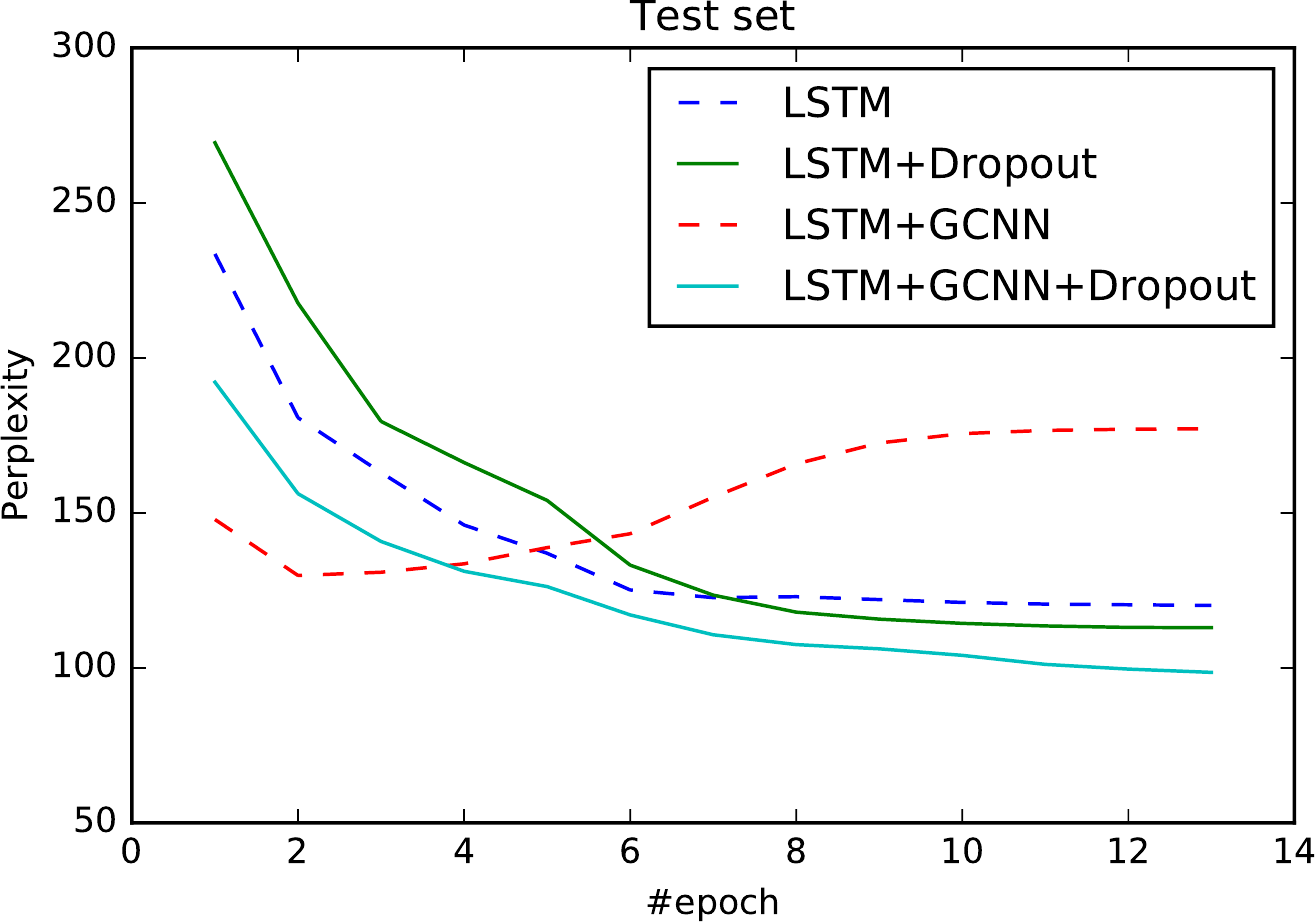}}
		\caption{Learning dynamic of LSTM with and without graph structure and dropout regularization.}
	\label{fig3}
\end{figure}

\begin{table*}[t]
	\centering
	{\small
		\begin{tabular}{lcccc}
			\toprule
			Architecture & Representation & Parameters & Train Perplexity & Test Perplexity  \\
			\midrule
			\citet{zaremba2014recurrent} code\footnotemark[6] & embedding & 681,800 & 36.96 & 117.29 \\
			\citet{zaremba2014recurrent} code\footnotemark[6] & one-hot & 34,011,600 & 53.89 & 118.82 \\
			LSTM & embedding & 681,800 & 48.38 & 120.90 \\
			LSTM & one-hot & 34,011,600 & 54.41 & 120.16 \\
			LSTM, dropout & one-hot & 34,011,600 & 145.59 & 112.98 \\
			GCRN-M1 & one-hot & 42,011,602 & 18.49 & 177.14 \\
			GCRN-M1, dropout & one-hot & 42,011,602 & 114.29 & {\bf 98.67} \\
			\bottomrule
		\end{tabular}
	}
	\caption{Comparison of models in terms of perplexity. \cite{zaremba2014recurrent} code$^6$ is ran as benchmark algorithm. The original \cite{zaremba2014recurrent} code used as input representation for $x_t$ the 200-dim embedding representation of words, computed here by the gensim library$^5$. As our model runs on the 10,000-dim one-hot representation of words, we also ran \cite{zaremba2014recurrent} code on this representation. We re-implemented \cite{zaremba2014recurrent} code with the same architecture and hyperparameters. We remind that GCRN-M1 refers to GCRN Model 1 defined in \eqnref{lstm_graph_v1}.} 
	\label{tab1}
\end{table*}

Table \ref{tab1} reports the final train and test perplexity values for each
investigated model and Figure \ref{fig3} plots the perplexity value vs. the
number of epochs for the train and test sets with and without dropout
regularization. Numerical experiments show:
\begin{enumerate}
\item Given the same experimental conditions in terms of architecture and {\it
	no} dropout regularization, the standalone model of LSTM is more accurate
	than LSTM using the spatial graph information ($120.16$ vs. $177.14$),
	extracted by graph CNN with the GCRN architecture of Model 1, Eq.
	\eqnref{lstm_graph_v1}. 
\item However, using dropout regularization, the graph LSTM model overcomes the
	standalone LSTM with perplexity values $98.67$ vs. $112.98$. 
\item The use of spatial graph information found by graph CNN speeds up the
	learning process, and overfits the training dataset in the absence of
	dropout regularization. The graph structure likely acts a constraint on the
	learning system that is forced to move in the space of language topics.
\item We performed the same experiments with LSTM and Model 2 defined in
	\eqnref{lstm_graph_v2}. Model 1 significantly outperformed Model 2, and
	Model 2 did worse than standalone LSTM. This bad performance may be the
	result of the large increase of dimensionality in Model 2, as the dimension
	of the hidden and cell states changes from 200 to 10,000, the size of the
	vocabulary. A solution would be to downsize the data dimensionality, as
	done in \cite{convlstm} in the case of image data.
\end{enumerate}

\footnotetext[6]{\url{https://github.com/tensorflow/tensorflow/blob/master/tensorflow/models/rnn/ptb/ptb\_word\_lm.py}}

\begin{figure}[t]
	\centering
	\subfigure{\includegraphics[height=5cm]{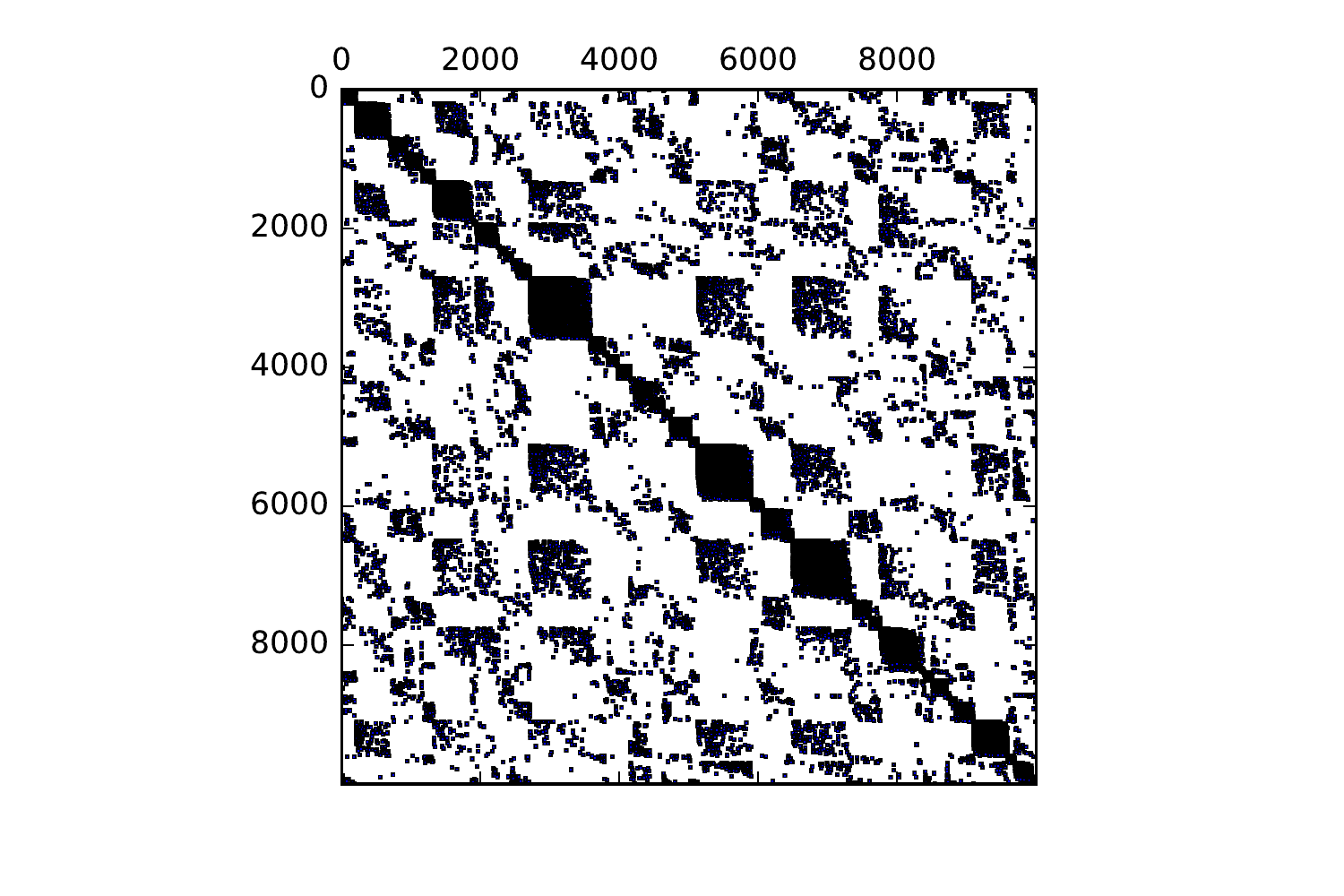}}
	\hspace{0.5cm}
	\subfigure{\includegraphics[height=5cm]{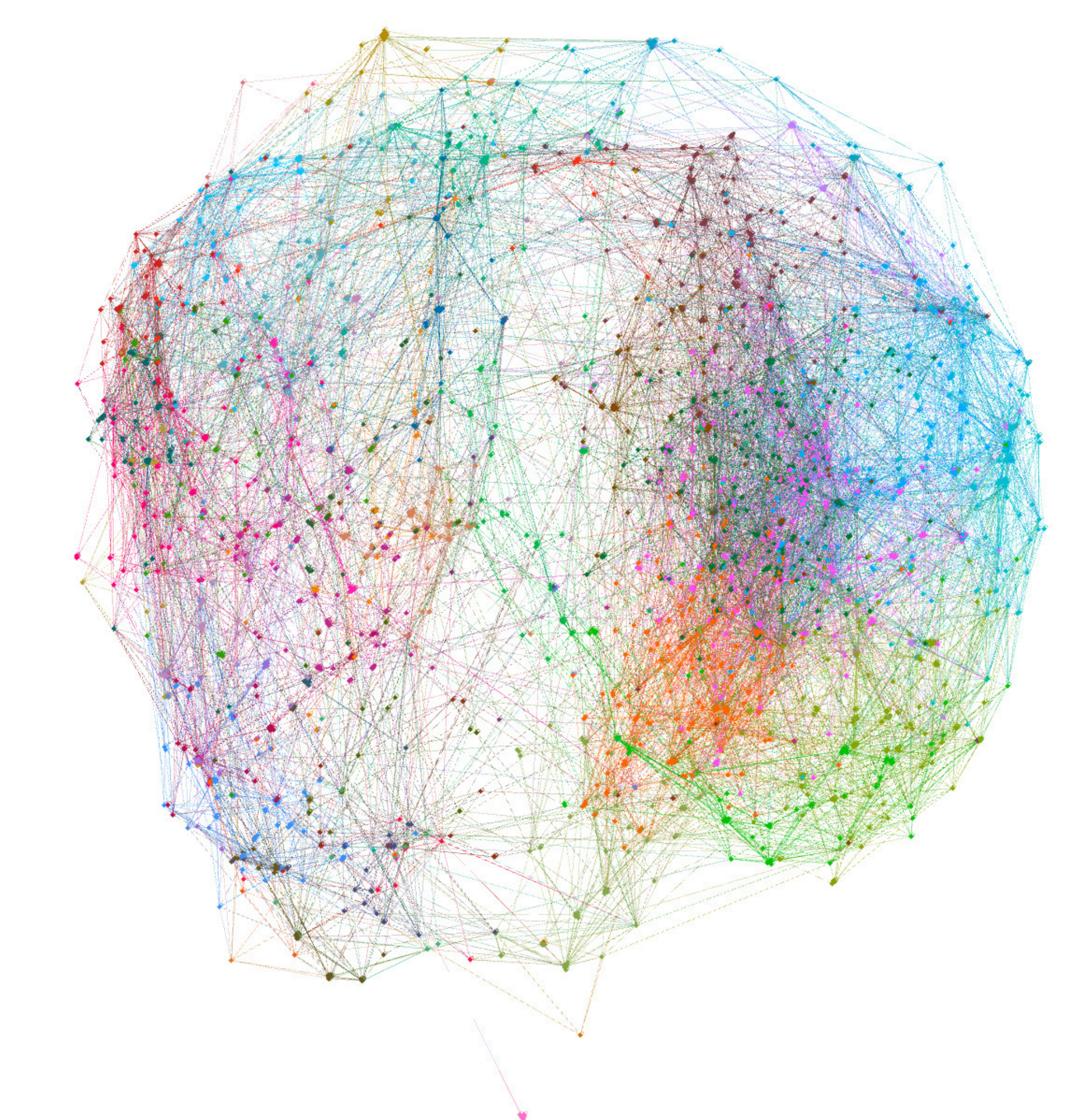}}
	\vspace{-0.6cm}
	\caption{Left: adjacency matrix of word embeddings. Right: 3D visualization of words' structure.}
	\label{fig2}
\end{figure}

\section{Conclusion and Future Work}

This work aims at learning spatio-temporal structures from graph-structured and
time-varying data. In this context, the main challenge is to identify the best
possible architecture that combines simultaneously recurrent neural networks
like vanilla RNN, LSTM or GRU with convolutional neural networks for
graph-structured data. We have investigated here two architectures, one using a
stack of CNN and RNN (Model 1), and one using convLSTM that considers
convolutions instead of fully connected operations in the RNN definition (Model
2). We have then considered two applications: video prediction and natural
language modeling. Model 2 has shown good performances in the case of video
prediction, by improving the results of \citet{convlstm}. Model 1 has also
provided promising performances in the case of language modeling, particularly
in terms of learning speed. It has been shown that (i) isotropic filters, maybe
surprisingly, can outperform classical 2D filters on images while requiring
much less parameters, and (ii) that graphs coupled with graph CNN and RNN are a
versatile way of introducing and exploiting side-information, e.g. the semantic
of words, by structuring a data matrix.

Future work will investigate applications to data naturally structured as
dynamic graph signals, for instance fMRI and sensor networks. The graph CNN
model we have used is rotationally-invariant and such spatial property seems
quite attractive in real situations where motion is beyond translation. We will
also investigate how to benefit of the fast learning property of our system to
speed up language modeling models. Eventually, it will be interesting to
analyze the underlying dynamical property of generic RNN architectures in the
case of graphs. Graph structures may introduce stability to RNN systems, and
prevent them to express unstable dynamic behaviors.


\section*{Acknowledgment}

This research was supported in part by the European Union's H2020 Framework
Programme (H2020-MSCA-ITN-2014) under grant No. 642685 MacSeNet, and Nvidia
equipment grant.

{
	\small
	\bibliography{iclr2017}
	\bibliographystyle{iclr2017}
}

\end{document}